\documentclass[a4paper,12pt]{article}

\usepackage{times}
\usepackage{ocg}
\usepackage{epsfig}

\title{On the Scalability of the Answer Extraction System ``ExtrAns''}

\author{Diego Moll\'a \and Michael Hess}

\renewcommand\refsectioning\relax

\begin{document}

\maketitle
\thispagestyle{empty} 

There have been many attempts in the history of Information Retrieval
(IR) to add some linguistic capabilites to standard IR systems in
order to improve their performance (mainly, their
precision).\footnote{This research is funded by the Swiss
  National Science Foundation, project nr. 12-53704.98} These
attempts have not been very successful so far, at least not in the
standard IR settings (cf.~\cite{Strzalkowski:TREC5}). The two main
reasons are the (related but not identical) problems of \emph{data
  volume} and of \emph{scalability}. First, the volume of data
typically processed by IR systems is so large that the use of more
than a few isolated linguistic components seemed out of the question,
and linguistic components do not work well in isolation. Second, NLP
systems that work reasonably well in small scale laboratory contexts
will often not scale up to real world domains like those for which IR
is standardly used. Both of these points seem to all but rule out the
use of full-fledged NLP methods in standard text retrieval
applications.

For some specific applications, however, high recall and precision are
even more crucial than in IR yet the volumes of data to process are
much smaller. These applications include interfaces to
machine-readable technical manuals, on-line help systems for complex
software (such as operating systems), help desk systems in large
organisations, and public inquiry systems accessible over the
Internet. In all these applications the document collections to be
accessed are just a few hundred megabytes in size at most.  The users
of these applications do not want a set of complete documents, each
one possibly dozens of pages long, as in standard IR.  What they want
is a few highly specific answers to their highly specific queries. In
other words, in such applications the user needs a system that locates
those exact \emph{phrases} in the documents which contain the explicit
answers to their queries. This is what an \emph{answer extraction
  system} is supposed to do, and it will require the use of linguistic
knowledge if it is to succeed. Note that answer extraction systems are
not meant to \emph{infer} answers from implicit information contained
in the documents (as is the idea of full-fledged text understanding
systems). All they should do is retrieve phrase-sized passages of text
containing an \emph{explicit} answer to a query, if there is one. This
is the task called, a bit confusingly, ``question answering'' in
TREC-8 (~\cite{TREC-8}).

We are currently developing such an answer extraction system for the
online Unix manpages. The system, ExtrAns, uses NLP as its core
technology to achieve the needed performance in terms of very high
precision and recall for highly specific
queries~\cite{Molla:Aug:1998,Berri:1998}.

\begin{figure}[htbp]
  \vspace{1em}
    \begin{center} 
        \epsfig{file=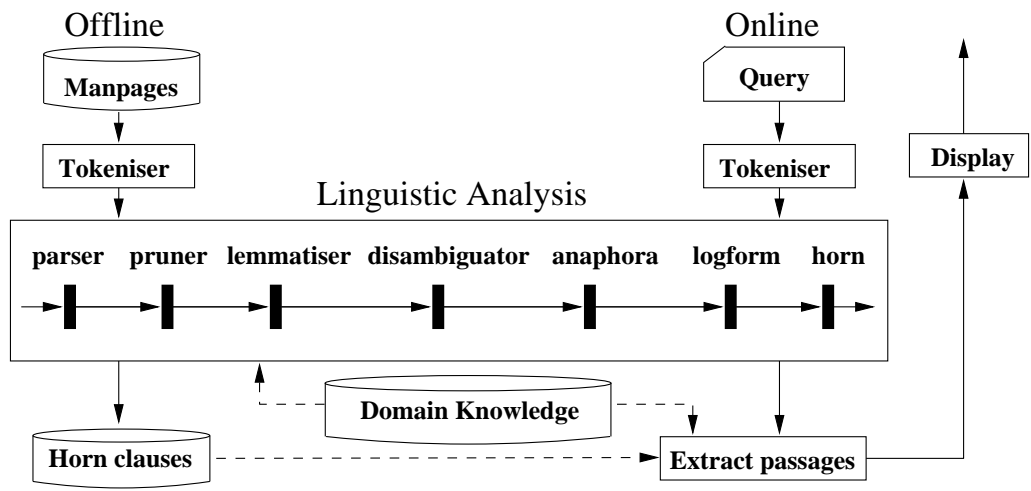,width=0.8\textwidth}
        \caption{Structure of ExtrAns} 
        \label{fig:extrans} 
    \end{center} 
\end{figure}

The overall structure of ExtrAns is shown in figure~\ref{fig:extrans}.
At indexing time, i.e. \emph{offline}, the manpages are subjected to a
full syntactic analysis, integrating the system developed
by~\cite{Sleator:1993}. We then weed out obvious wrong readings of the
input sentences by using a set of specific rules. The words are then
converted to their base form by using the lemmatiser included
in~\cite{Gaizauskas:1996}. More difficult ambiguous readings are
deleted by adapting and enhancing~\cite{Brill:1994}, a corpus-based
disambiguator. Next, intra-sentential pronominal references are
resolved, following the algorithm suggested by~\cite{Lappin:1994}.
Finally, logical forms are derived, transformed into Horn clauses, and
stored in a database.

\begin{figure}[htbp]
  \vspace{1em}
  \begin{center}
    \epsfig{file=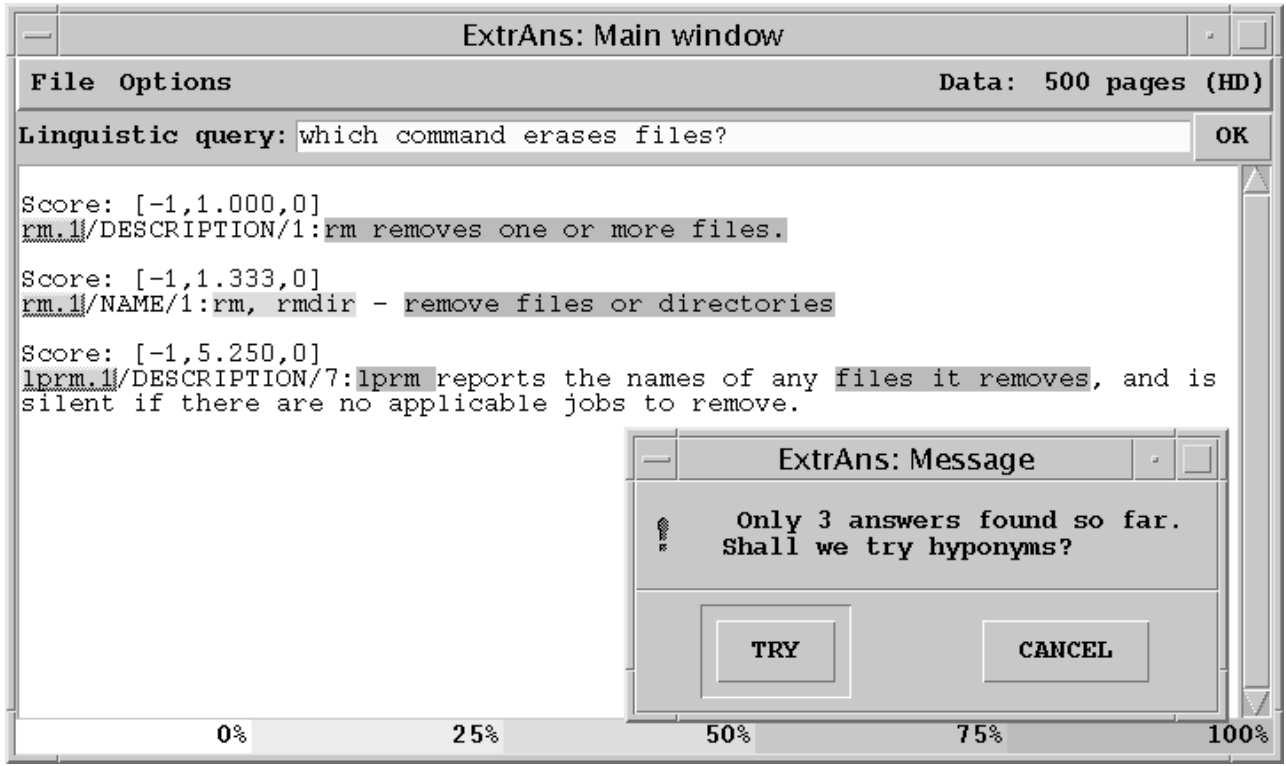,width=\textwidth}
    \caption{A screen shot of ExtrAns (monochrome rendering)}
    \label{fig:screenshot}
  \end{center}
\end{figure}

The user can then freely query the system by asking questions in plain
English. The logical form of the query is computed, \emph{online}, in
the same way as described above, and the system tries to prove the
query over the database. If successful, the corresponding sentences
are retrieved and displayed, with those phrases that explicitly answer
the user query highlighted according to their relevance to the query,
where relevance is a factor of (un)ambiguity~\cite{Molla:Aug:1998}.
Figure~\ref{fig:logforms} shows the logical forms of a query and an
answer (see~\cite{Molla:Aug:1998} for more details about logical
forms), and figure~\ref{fig:screenshot} shows a screen shot of the
output of ExtrAns for the same query.

\begin{figure}[htbp]
  \vspace{1em}
  \begin{center}
    \renewcommand{\baselinestretch}{1}
    \small
    \begin{quote}
    \emph{which command erases files?}\\
      \texttt{object(s\_command,A,B), evt(s\_remove,C,[B,D]),
        object(s\_file,E,D)}

    \emph{rm removes one or more files}\\
      \texttt{holds(v\_e2), object(rm,v\_o\_a1,v\_x1),
        object(s\_command,v\_o\_a2,v\_x1),
        evt(s\_remove,v\_e2,[v\_x1,v\_x6]),
        object(s\_file,v\_o\_a3,v\_x6)}
    \end{quote}
    \caption{The logical form of a user query and an answer (simplified to ease readability)}
    \label{fig:logforms}
  \end{center}
\end{figure}

In order to test scalability, we started with a small subset of 30
Unix manpages as development set and extended the document basis, in a
second step, to a test set of over 500 manpages.  We tuned the system
so as to be able to cope with the larger volumes of data but did not
change any of the linguistic components (we only extended the lexicon
to increase the accuracy of the parses, although the parser
used~\cite{Sleator:1993} does handle unknown words). A brief
description of the modifications follows:

\vspace{-2ex}

\begin{enumerate}
\item The set of Horn clauses was stored in an external database, since a
  larger size of the data would not fit in the RAM memory. An obvious
  consequence of this decision is that the speed of retrieval would
  degrade, but we solved this in the second set of modifications:
\item The database of Horn clauses was divided into a set of
  databases, one for each manpage, and a pre-selection step was added
  so that the query is run only over those manpages that contain all
  of the terms used in the logical form of the query. In this fashion,
  only those manpages that are likely to contain the answer to the
  question are examined.
\end{enumerate}

\vspace{-2ex}

We found that the system retrieved the same sentences as before and
that response times, after performing the modifications, were shorter,
even if the number of manpages treated was considerably larger. In
table~\ref{tab:times} we can also see that the relative increase of
response time is well lower than the ratio of the sizes of the data of
both sets of manpages ($272\hbox{Kbytes}/20\hbox{Kbytes}=13.6$).
The original system thus turned out to be perfectly scalable, contrary
to what is normally assumed to be the case for NLP-based retrieval
systems.

\begin{table}[htbp]
  \vspace{1em}
  \begin{center}
   \renewcommand{\baselinestretch}{1}
   \small
    \begin{tabular}{l|rr|r|r}
      & \multicolumn{2}{c|}{\textbf{30 manpages}} &
      \textbf{500 manpages}\\ 
      \textbf{Sentence} & \textbf{Internal} (a) & \textbf{External} (b) &
      \textbf{External} (c) & (c) / (b)\\
      \hline
      \emph{which command copies files?} & 2980 & 584 & 3206 & 5.48\\
      \emph{how can I create a directory?} & 7412 & 1270 & 3374 & 2.65\\
      \emph{which command removes directories?} & 4632 & 416 & 1080 & 2.59\\
      \emph{how can a file be removed?} & 5110 & 936 & 4584 & 4.89\\
      \emph{can I remove some columns from a text file?} & 5750 & 316
      & 384 & 1.21\\
      \emph{what is ipcrm?} & 258 & 194 & 246 & 1.26\\
      \emph{which command erases files?} & 7430 & 794 & 3986 & 5.02\\
    \end{tabular}
    \caption{Response time in \emph{ms.} from two sets of
      manpages, on a 167-MHz UltraSparc machine. The set of 30
      manpages was stored in an internal (a) and an external (b)
      database. The set of 500 manpages was stored in an external
      database (c).}
    \label{tab:times}
  \end{center}
\end{table}

The idea of preselecting manpages is connected with another
possibility that we are considering. We could include a standard
Information Retrieval (IR) module specifically tuned-up to give
results with high recall. The IR module would provide ExtrAns with a
reduced set of data, and ExtrAns would use its linguistically-aware
techniques to further reduce the amount of data, so that eventually
the user would get the wanted answers with a high index of recall and
precision. By combining standard IR techniques with a system such as
ExtrAns it would be possible to find answers to queries over data in
the size scale of gigabytes.

By combining available linguistic resources and implementing only a
few modules from scratch, we have been able to put together a system
with full linguistic analysis in a relatively short period of time (4
man-years). Table~\ref{tab:times} shows that the increase in time is
not a big issue when scaling the system up from 30 to 500 documents.
The main result of our experiment is thus that the use of existing NLP
techniques allows us to implement answer extraction systems whose
performance is perfectly acceptable, even in their laboratory
versions, and that such systems do scale up to practical dimensions.

\bibliographystyle{ocg}

{
\renewcommand\em\relax
\bibliography{database}
}
\end{document}